\documentclass[10pt,twocolumn,letterpaper]{article}

\usepackage{cvpr}              
\usepackage{amsmath} 
\usepackage{caption}
\usepackage{capt-of}
\usepackage{bbding}
\usepackage{amssymb}
\usepackage{pifont}
\usepackage{wrapfig}
\usepackage{hyperref}
\usepackage{graphicx}
\usepackage{subcaption}
\usepackage{hyperref}
\usepackage{natbib}
\setcitestyle{numbers,square}
\usepackage[ruled,vlined]{algorithm2e}
\usepackage[subrefformat=parens,labelformat=simple]{subcaption} 

\hypersetup{colorlinks,linkcolor={red},citecolor={green},urlcolor={magenta}} 

\usepackage[dvipsnames]{xcolor}

\definecolor{cvprblue}{rgb}{0.21,0.49,0.74}

\def\paperID{3700} 
\def\confName{CVPR}
\def\confYear{2025}

\title{PLA4D: Pixel-Level Alignments for Text-to-4D
Gaussian Splatting}

\author{Qiaowei Miao\\
Zhejiang University\\
Hangzhou, China\\
{\tt\small qiaoweimiao@zju.edu.cn}
\and
Jinsheng Quan\\
Zhejiang University\\
Hangzhou, China\\
{\tt\small jinshengquancv@gmail.com}
\and
Kehan Li\\
Zhejiang University\\
Hangzhou, China\\
{\tt\small kehanli@zju.edu.cn}
\and
Yawei Luo*\\
Zhejiang University\\
Hangzhou, China\\
{\tt\small yaweiluo@zju.edu.cn}
}

\begin{document}
\maketitle
\begin{abstract}

Previous text-to-4D methods have leveraged multiple Score Distillation Sampling (SDS) techniques, combining motion priors from video-based diffusion models (DMs) with geometric priors from multiview DMs to implicitly guide 4D renderings. However, differences in these priors result in conflicting gradient directions during optimization, causing trade-offs between motion fidelity and geometry accuracy, and requiring substantial optimization time to reconcile the models. In this paper, we introduce \textbf{P}ixel-\textbf{L}evel \textbf{A}lignment for text-driven \textbf{4D} Gaussian splatting (PLA4D) to resolve this motion-geometry conflict. PLA4D provides an anchor reference, i.e., text-generated video, to align the rendering process conditioned by different DMs in pixel space. For static alignment, our approach introduces a focal alignment method and Gaussian-Mesh contrastive learning to iteratively adjust focal lengths and provide explicit geometric priors at each timestep. At the dynamic level, a motion alignment technique and T-MV refinement method are employed to enforce both pose alignment and motion continuity across unknown viewpoints, ensuring intrinsic geometric consistency across views. With such pixel-level multi-DM alignment, our PLA4D framework is able to generate 4D objects with superior geometric, motion, and semantic consistency. Fully implemented with open-source tools, PLA4D offers an efficient and accessible solution for high-quality 4D digital content creation with significantly reduced generation time.

\end{abstract}    
\section{Introduction}

\begin{figure}
    \centering
    \includegraphics[width=0.9\linewidth]{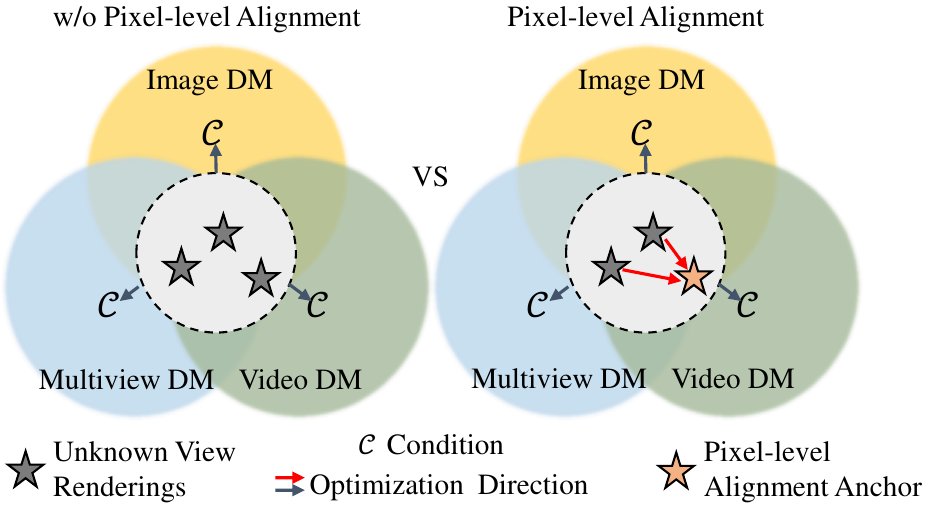}
    \caption{Without offering an anchor reference in pixel space, multiple SDS align each rendering to their respective priors, which may not be consistent across different diffusion model priors, requiring significant time for reconciliation to generate a 4D result. With the anchor reference in pixel space, however, each SDS can optimize the 4D geometry and motion representation according to its respective prior more effectively.}
    \vspace{-10pt}
    \label{fig:intro}
\end{figure}

Text-to-4D content generation has significant potential in applications ranging from game production to autonomous driving. However, this task remains challenging due to the need to generate high-quality geometry and textures, alongside coherent object animations aligned with textual prompts. Existing methods in text-to-4D synthesis, such as MAV3D~\cite{make-a-video-3d} and 4D-fy~\cite{4d-fy}, often employ Neural Radiance Fields (NeRF)~\cite{Nerf}. MAV3D achieves text-to-4D generation by distilling text-to-video diffusion models (DMs) onto a Hexplane\cite{cao2023hexplane}, while 4D-fy utilizes multiple pre-trained DMs with hybrid score distillation sampling (SDS) to generate compelling 4D content. Recent approaches, like AYG~\cite{AYG}, leverage 3D Gaussians deformed by a neural network and incorporate multiple SDS modules from text-to-image, text-to-multiview, and text-to-video DMs~\cite{mvdream,Alignyourlatents} to guide geometry and motion generation.

A commonality among the above methods is their heavy reliance on the SDS of multiple DMs to provide priors for guiding the generation of geometry and motion. However, the reliance on SDS-based methods brings considerable challenges. As shown in Fig.~\ref{fig:intro}, the goal of SDS can be viewed as leveraging the priors from DMs to implicitly align rendered images with conditions $\mathcal{C}$. However, due to the different source datasets each DM is pre-trained on, even with the same condition, the results generated by different DMs vary. This discrepancy can lead to conflicts when multiple DMs are jointly optimized using SDS, resulting in two primary issues: (1) Motion-geometry trade-off. When video DMs and multiview DMs have conflicting optimization targets, it becomes challenging to generate 4D outputs that balance both motion and geometry. Since the SDS implicitly aligns rendered images with the condition, we cannot easily adjust the scale of their losses for a better motion or a better geometry. (2) Excessive optimization time. When conflicts arise between multiple SDS losses, a substantial amount of time is required to balance these conflicting objectives, which is one of the main reasons for the time-consuming nature of current methods.

In this paper, we introduce a novel framework for text-to-4D content creation, dubbed \textbf{PLA4D} (\textbf{P}ixel-\textbf{L}evel \textbf{A}lignments for Text-to-\textbf{4D} Gaussian Splatting), which generates 4D objects with video-like smooth motion from text in exceptionally short time. Our core idea is to shift from implicit latent-level alignments to explicit pixel-level alignment. By using text-generated video as an anchor, we ensure that rendered images are simultaneously aligned with both prompt and pixel representations across the priors of multiple DMs.   To achieve this, we approach the problem at static and dynamic levels, with each level incorporating several novel modules. In the static alignment module, we introduce the Focal Alignment module to estimate the corresponding focal length of each generated frame, which generates a reference mesh corresponding to the video frame by an image-to-mesh diffusion model. It then estimates the focal length of each generated frame by calculating the similarity between mesh renderings and video frames at different focal lengths. With the correct focal length, the current frame can accurately supervise the primary viewpoint rendering of 4D at the corresponding timestep. Consequently, we introduce Gaussian-Mesh Contrastive Learning, which utilizes the mesh during the focal length alignment to provide geometric supervision, thus maintaining geometric consistency for unknown views.

In the dynamic alignment module,  we need to consider both temporal and multiview consistency. We guide the motion of 4D outputs to align with the anchor video, transferring the motion guidance from the video to the 4D target. Simultaneously, we ensure coherent motion across different viewpoints. To achieve motion continuity, we randomly select a viewpoint for rendering multiple timesteps and align this with the text conditions under the guidance of a video DM that generates the anchor video. This approach enables the geometry and texture learned from static alignment to smoothly extend across the temporal dimension. Besides, the motion performance of 4D objects in unknown views can align with the anchor video.
To further reinforce consistency in unseen viewpoints, we randomly choose a timestep and render images across multiple views, enhancing their consistency with the corresponding frame of anchor video under the guidance of a multiview DM. Through the combined effects of static alignment and dynamic alignment modules, PLA4D enables text-driven-generated 4D objects to have geometric consistency, smooth and semantically aligned motion, and minimal time overhead.

PLA4D can generate a wide range of dynamic objects rapidly, producing diverse, vivid, and intricate details while maintaining geometry consistency, as shown in Fig.~\ref{fig:abs}. In summary, our contributions are as follows:
\begin{itemize}
\item We present a novel text-driven 4D generation framework that leverages explicit anchor reference, i.e., text-generated video, to align the rendering process conditioned by different DMs in pixel space, eliminating the optimization conflicts of different DMs.
\item We propose focal alignment and Gaussian-Mesh contrastive learning, which automatically finds the best focal parameters corresponding to reference pixels and explicitly provides geometry guidance for 4D.
\item We propose a motion alignment method and Time-Multiview refinement modules to optimize 4D, ensuring video-like, large motions aligned with textual semantics.
\item PLA4D achieves remarkable performance, generating 4D objects with fine textures, accurate geometry, and coherent motion in significantly less time.
\end{itemize}

\section{Related work}

\textbf{3D Generation.}
Recent advancements in DMs within 2D domains have sparked significant interest in exploring 3D generative modeling~\cite{Fantasia3d,It3d, Nerdi, MetaDreamer,Dreamtime, Noise-free, Luciddreamer,Magic3d, Att3d,Dreamcraft3d,Textmesh, Scorejacobian, survey} for content generation. Under given control conditions (\emph{e.g.}, text prompt or single image), some efforts~\cite{mvdream,Zero-1-to-3,Zero123++,one-2-3-45,Syncdreamer} are made to extend 2D DMs from single-view images to multiview images to seamlessly integrate with different 3D representation methods (e.g, Nerf~\cite{Nerf}, Mesh~\cite{Mesh}, and 3D Gaussian~\cite{3DGS}).
However,  due to the uncertainty of the diffusion model's denoise process, the multiview consistency and corresponding camera poses of generated images are not guaranteed, leading to artifacts and texture ambiguity in the generated 3D object.  
Further, some works~\cite{Dreamfusion,Prolificdreamer,wang2023score} apply SDS ~\cite{dreamgaussin} in latent space to extend the 2D DMs to guide 3D generation. Although such SDS-based methods can improve the textural of 3D representation, they frequently suffer from Janus-face problems due to the lack of comprehensive multiview knowledge.
Recently, some methods~\cite{Magic3d,3DGS,It3d,Att3d} have integrated the above two approaches, which use pre-trained multiview diffusion for SDS. The comprehensive multiview knowledge or 3D awareness hidden in the pre-trained model enhances the consistency of 3D representation, yet such SDS-based methods are time-consuming, needing hours to train.

\noindent \textbf{Video Generation.}
Video generation~\cite{Latent-shift, Alignyourlatents,ge2023preserve, Emuvideo, Animatediff, Imagenvideo, Text2video-zero, Stablevideo}, including text-to-video and image-to-video generation, has been getting more and more attention recently. The former, such as MAV~\cite{make-a-video}  and AYL~\cite{Alignyourlatents}, rely on large amounts of high-quality text-to-video data for training to deepen their understanding of verbs, enabling them to generate rich and creative sequences of coherent video frames. The latter~\cite{Animatediff, Stablevideo} infers subsequent actions of the target object based solely on a given initial frame image, which does not support flexible control over actions.

\noindent \textbf{4D  Generation.} At the current stage, 4D generation is influenced by various factors. (1) Representation Methods: Previous methods have mainly been based on NeRF~\cite{Dream-in-4D,4d-fy,make-a-video-3d}, where its multi-layer MLP architecture facilitates the generation of smooth 4D surfaces, but requires a significant amount of time for training. Recently, some methods~\cite{AYG,Stag4d} based on 4D GS have emerged. While training speeds have improved, guiding the motion of each Gaussian point to drive the 4D target raises higher requirements for motion guidance. (2) Motion Guidance Methods: Some previous methods~\cite{dreamgaussin4d,4dgen,Stag4d} used image-to-video models to accomplish the image-to-4D task. However, the generated motions do not support user manipulation, significantly limiting usability. Using text-to-video models for guidance is a better approach. But current methods, such as MAV3D~\cite{make-a-video-3d} and AYG~\cite{AYG}, rely on closed-source video models~\cite{Alignyourlatents,make-a-video}. 4D-fy~\cite{4d-fy} attempts to use the open-source video model and SDS~\cite{dreamgaussin} to distill motion priors, but our experimental results show that this can only provide very limited motion. (3) Training Duration: Current text-to-4D methods are trained directly from a random initialization state based on SDS. Due to inconsistent optimization objectives for each SDS, a substantial amount of time is required for compromise, leading to generation times that often take hours. To address these challenges, we propose PLA4D, which is based on 4D GS. It uses a text-to-video model to provide pixel-level motion guidance and generates 4D objects quickly with mesh geometry priors.

\section{Methodology}



\subsection{Preliminaries}\label{sec:pre}
\textbf{4D Gaussian Splatting} is derived from 3D GS~\cite{3DGS}  by extending it along the time dimension via another model, such as the deformation network. 3D Gaussian involves a collection of $N$ Gaussian points, each defined by four attributes: positions $\mu_{i}$,
covariances $\Sigma_{i}$, colors $\ell_{i}$, and opacities $\alpha_{i}$. 
A common approach to incorporating time is to add a deformation network that predicts the attributes of each Gaussian point at each timestep.
To render novel views images at time $\tau$, 4D Gaussians fix time parameter and reproject the 3D Gaussians onto a 2D image space, obtaining their projection positions $\mu$ and corresponding covariances $\hat{\Sigma}_{i}$.  
Point-based $\alpha$\text{-}blending rendering~\cite{EWA} is then applied to determine the color $\mathcal{C}(\mathrm{p})$ of image pixel $\mathrm{p}$ along a ray $r$:
\begin{align}
    \mathcal{C}(\mathrm{p}) &= \sum_{i\in N}\ell_{i}\eta_{i}\prod_{j=1}^{i=1}(1-\eta_{j}),  \\ 
    \eta_{i} &= \alpha_{i}\mathrm{exp} \left [-\frac{1}{2}(\mathrm{p}-\hat{\mu}_{i})^{T}\hat{\Sigma}_{i}(\mathrm{p}-\hat{\mu}_{i}) \right ],
\end{align}
where $j$ iterates over the points traversed by the ray $r$, $\ell_{i}$ and $\alpha_{i}$ donate the color and opacity of the i\text{-}th Gaussian. $\hat{\mu}_{i}$ is the projection of $\mu_{i}$ on 2D image plane. Within each moment, the deformation network predicts a variable for each Gaussian point’s attributes and adds it on them, thus driving the 4D object's motion across multiple times.

\begin{figure*}[t!]
    \centering
    \includegraphics[width=\linewidth]{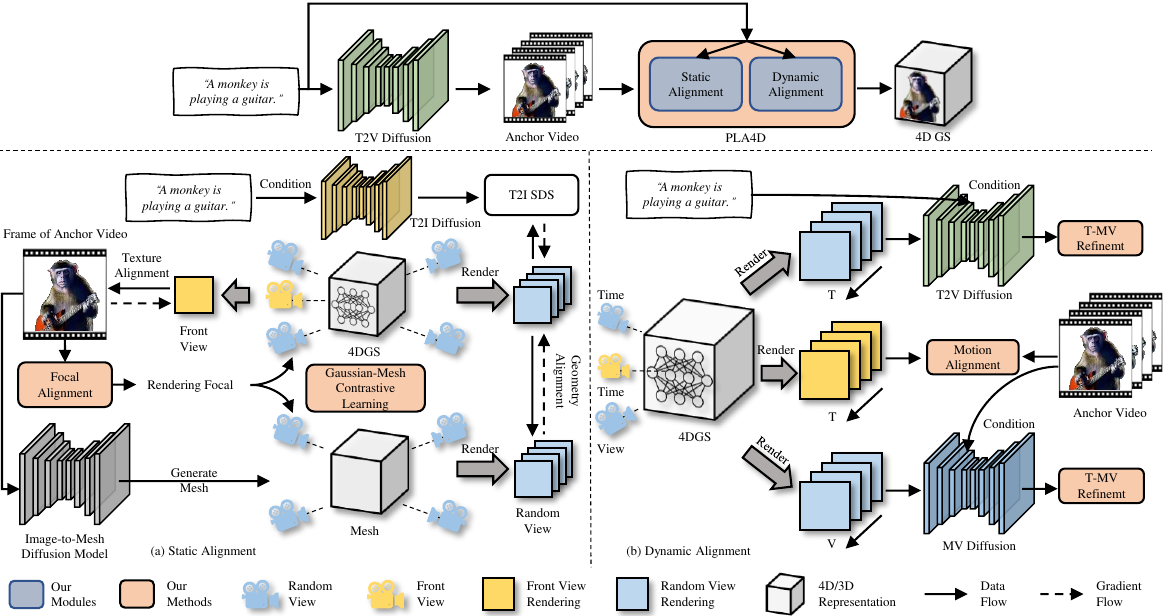}
    \caption{Pipeline of PLA4D, which leverages text as the condition and text-generated video as an anchor for 4D generation. (a) Static alignment: We propose focal alignment to search for the best focal length for 4D automatically. We also introduce Gaussian-Mesh Contrastive Learning to provide geometric information for 4D Gaussian in unknown views, explicitly leveraging the geometric priors of the mesh. (b) Dynamic alignment: Across multiple frames, we introduce motion alignment to guide the 4D object's motion following the anchor video. Furthermore, we propose Time-Multiview (T-MV) refinement to optimize the motion and quality of the 4D object's unknown viewpoints, using the prior and the condition of the model that generates the video.}
    \label{fig:overall}
    \vspace{-15pt}
\end{figure*}

\noindent \textbf{Score Distillation Sampling (SDS)} is widely used in 3D generation methods~\cite{mvdream, Dreamfusion, dreamgaussin4d, mvdream, Prolificdreamer}, which aligns the 3D generation process to the 2D DMs training process. 
In the training of 2D DMs,  sample noise $\epsilon$ from $q(\mathrm{x})$ and add it on the data $\mathrm{x}$ (\emph{e.g.}, images and videos) with $t$ times until $q(\mathrm{x}_{t})$ converges to a Gaussian prior distribution $\mathcal{N}(0,\textbf{I})$. The network $\phi$ is trained to predict the removal noise $\hat{\epsilon}$ for denoising and reconstructing data $\mathrm{x}$. 3D generation methods set the renderings got from 3D scene representation as data $\mathrm{x}$, and calculate the Mean Squared Error (MSE) to get SDS gradient~\cite{Dreamfusion}:
\begin{equation}
    \label{eq:sds_grad}
    \nabla_{\theta}\mathcal{L}_{\mathrm{SDS}}(\mathrm{x}=g(\theta))=\mathbb{E}_{t,\epsilon} \left [ w(t)(\hat{\epsilon}_{\phi}(\mathrm{z}, v, t) - \epsilon) \frac{\partial \mathrm{x}}{\partial \theta} \right ],
\end{equation}
where $t$ is the timestamp of denoising process, and $w(t)$ is time-dependent weights. $\mathrm{z}$ represents the latent of $\mathrm{x}$ needing to denoise. $v$ indicates the given conditions, such as text prompts and images. The SDS gradients are then backpropagated through the differentiable rendering process $g$ into the 3D representation and update its parameters $\theta$.

Previous 4D generation tasks use multiple DMs to obtain prior information about multiview and motion via SDS. For example, in the absence of pixel-level alignment targets, the SDS gradient directions of each DM might conflict with each other. Consider a simple case where only video and MV DMs are used, the SDS gradient is:
\begin{multline}
    \nabla_{\theta}\mathcal{L}_{\mathrm{SDS}}(\mathrm{x}=g(\theta))= \\
    \mathbb{E}_{t,\epsilon} \left [ w(t)(\hat{\epsilon}_{\phi_{V}}(\mathrm{z}, v, t)+\hat{\epsilon}_{\phi_{MV}}(\mathrm{z}, v, t) - 2\epsilon) \frac{\partial \mathrm{x}}{\partial \theta} \right ],
\end{multline}
the overall SDS gradient descent direction should be the vector sum of the multi\text{-}view DM SDS and video DM SDS gradient descents. However, this is not stable. When the two gradient directions are opposite, the 4D model will be caught in a dilemma, requiring extensive optimization time to find a local optimum. As the number of DMs increases, this issue becomes more pronounced.

\subsection{Pipeline of PLA4D} \label{sec:Overall}

PLA4D introduces static alignment and dynamic alignment modules to achieve text-driven 4D generation, leveraging multiple DMs, including T2V DM, I2MV DM, and T2I DM, as illustrated in Fig.~\ref{fig:overall}. T2V DM is used to generate an anchor video and refine the motion of the 4D object, while I2MV DM refines the geometry of the 4D object from unknown views. T2I DM bridges the gap between the anchor video and the pixel-level geometry priors provided by the mesh. Inspired by DreamGaussian4D~\cite{dreamgaussin4d}, we combine 3D Gaussians with a deformation network to support 4D generation. Initially, we use an open-source T2V DM to generate an anchor video and use Eq.\ref{eq:intra} of the static alignment to get 3D Gaussian as initialization for 4D Gaussian. Next, we apply Eq.\ref{eq:intra} of the static alignment and Eq.~\ref{eq:inter} of the dynamic alignment modules to optimize the deformation network.

\subsection{Static Alignment Module} \label{sec:3D}

\begin{figure}[htbp]
\centering
  \begin{minipage}[b]{0.44\linewidth}{
    \includegraphics[width=1\linewidth]{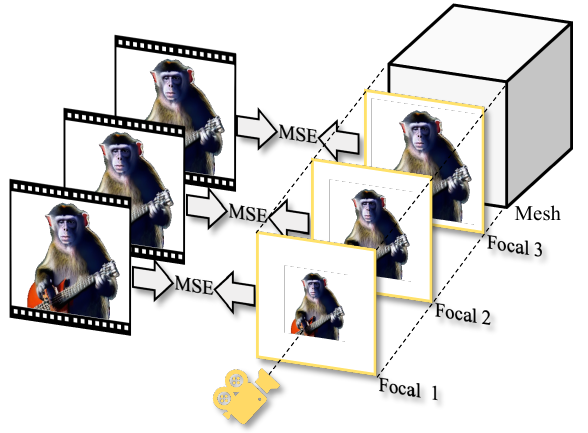}
    \subcaption{Focal Alignment}\label{fig:3d1}
    }
  \end{minipage}
  \begin{minipage}[b]{0.55\linewidth}{
    \includegraphics[width=1\linewidth]{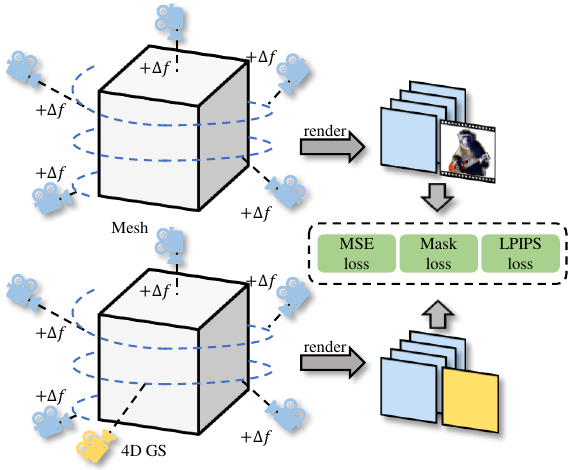}
    \subcaption{Gaussian-Mesh Contrastive Learning}\label{fig:3d2}
    }
  \end{minipage}
    \caption{Focal alignment and Gaussian-Mesh contrastive learning. (a) We render multiple front-view images and calculate the MSE with the first frame for searching the matched focal. (b) We collect two sets of images: one of 4D Gaussians and another of mesh renderings, both captured using the same random camera poses. We include the first frame of the anchor video and the front-view renderings in these two sets. Then, we calculate the MSE loss, Mask loss, and LPIPS loss between the corresponding images.}
    \label{fig:3d}
    \vspace{-10pt}
\end{figure}


\textbf{Focal Alignment for Texture Alignment.}
PLA4D aims to use text-generated video as the pixel-level alignment anchor for 4D generation, which needs a matched focal for frames.
However, anchor frames' focals are unknown. Therefore, we propose focal alignment to search for the matched focal $f$.
Specifically, we start with the video synthesis. Given the text prompt $v$, PLA4D applies a text-to-video DM $G_{\mathrm{vid}}$ to create a video $\left \{ I_\mathrm{vid}^{t} \right \}_{\mathcal{T}} = G_{\mathrm{vid}}(\epsilon; v)$ with $\mathcal{T}$ frames. $\epsilon$ is a random noise. Because the view angles in the anchor frames are relatively fixed, we set the video's view $c$ as the 4D object's front perspective.
Next, at the beginning of each timestep $t$, we need to fix the time parameter of 4D Gaussian and compare its front view rendering and $I_\mathrm{vid}^{t}$ to search $f'$, as shown in Fig.~\ref{fig:3d1}. 
Hence, we introduce CRM~\cite{Crm}, an image-to-mesh feed-forward 3D generation model, to generate a mesh $\psi_{t}$ based on $I_\mathrm{vid}^{t}$. 
We render $\psi_{t}$'s front-view images $\left \{\mathrm{x}_{\psi_{t}} \right \}_{M}$ with $M$ different focals iterated from $f'+\Delta f_{min}$  to $f'+\Delta f_{max}$, where $f'$ is an initial focal length.
We calculate the Mean Squared Error (MSE) between $I_\mathrm{vid}^{t}$ and $\left \{\mathrm{x}_{\psi_{t}} \right \}_{M}$ for searching the matched focal $f'$:
\begin{equation}
    f =\mathop{\arg\min}\limits_{f'}\sum_{H,W}||\mathrm{x}_{\psi_{t}}^{f'} - I_\mathrm{vid}^{t}||^{2}_{2}. 
\end{equation}

At each timestep, with the corresponding focal $f$, we propose Gaussian-Mesh contrastive learning to align the front-view 4D Gaussian renderings to the frames to achieve texture alignment, which is composed of three losses: (I)
$\mathcal{L}_{\mathrm{MSE}}$ for aligning the pixel-level similarity, (II) $\mathcal{L}_{\mathrm{Mask}}$ for reducing the floaters, and (III) $\mathcal{L}_{\mathrm{LPIPS}}$ for enhancing the visual perceptual perception.
In particular, we use the MSE loss between front view $c$ rendering $\mathrm{x}_{\theta}$ of 4D Gaussians  and $ I_\mathrm{vid}^{t}$  as follows:
\begin{equation}
    \mathcal{L}_{\mathrm{MSE}}(\mathrm{x}_{\theta_{t}}^{c} , I_\mathrm{vid}^{t}) =  \sum_{H,W} || \mathrm{x}_{\theta_{t}}^{c} - I_\mathrm{vid}^{t} ||_{2}^{2}.
\end{equation}
Besides, to reduce the floaters, we also use the transparent output $\alpha$ of 4D Gaussians as the mask and calculate the mask loss:
\begin{equation}
    \mathcal{L}_{\mathrm{Mask}}(\mathrm{x}_{\theta_{t}}^{c} , I_\mathrm{vid}^{t}) = \sum_{H,W} || \alpha_{\theta_{t}}^{c} - \alpha_\mathrm{vid}^{t} ||_{2}^{2},
\end{equation}
where $\alpha_\mathrm{vid}^{t}$ is the alpha channel of $ I_\mathrm{vid}^{t}$. Besides, we introduce Learned Perceptual Image Patch Similarity (LPIPS)~\cite{LPIPS}, which is a metric used to measure perceptual differences between images. We apply LPIPS  loss between $\mathrm{x}_{\theta_{t}}^{c}$ and $I_\mathrm{vid}^{t}$ to enhance the visual quality of textures. 
$\mathcal{L}_{\mathrm{LPIPS}}$ needs an encoder (\emph{i.e.}, VGG~\cite{vgg}) to extract feature stack from $l$ layers and unit-normalize in the channel dimension, and calculate the MSE between features extracted from each layer:
\begin{equation}
    \mathcal{L}_{\mathrm{LPIPS}}(\mathrm{x}_{\theta_{t}}^{c} , I_\mathrm{vid}^{t}) = \sum_{l}\frac{1}{H_{l}W_{l}}\sum_{H_{l}W_{l}}||\mathrm{z}_{\theta_{t}}^{c} - \mathrm{z}_\mathrm{vid}^{t} ||^{2}_{2}.
\end{equation}
Now, we can get the texture alignment loss $\mathcal{L}_{\mathrm{TA}}$:
\begin{align}
    \mathcal{L}_{\mathrm{TA}}& =   \mathcal{L}_{\mathrm{MSE}}(\mathrm{x}_{\theta_{t}}^{c} , I_\mathrm{vid}^{t}) \notag \\ & +   \mathcal{L}_{\mathrm{Mask}}(\mathrm{x}_{\theta_{t}}^{c} , I_\mathrm{vid}^{t}) +  \lambda  \mathcal{L}_{\mathrm{LPIPS}}(\mathrm{x}_{\theta_{t}}^{c} , I_\mathrm{vid}^{t}),
\end{align}
where  $\lambda$ is the scaling weight for balance.

\begin{figure*}[hpt]
    \centering
    \includegraphics[width=1\linewidth]{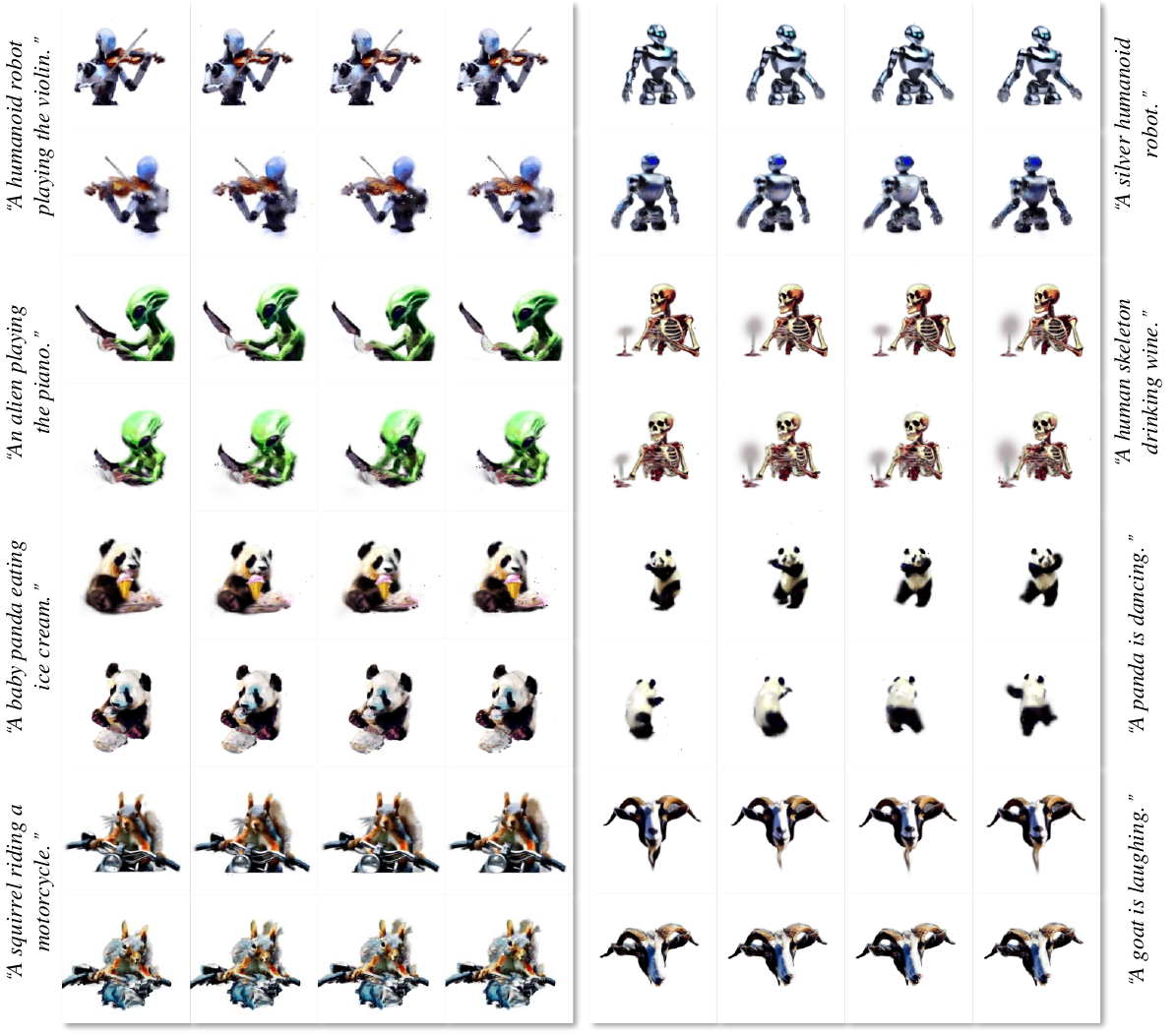}
    \caption{Visualization results of PLA4D. 
The 4D objects generated by PLA4D not only rigorously follow the semantics but also feature-rich dynamics and excellent geometric consistency. More importantly, PLA4D generates each sample in approximately 15 minutes. }
    \label{fig:show}
\end{figure*}

\noindent \textbf{Gaussian-Mesh Contrastive Learning for Geometry Alignment.}
Thanks to our focal alignment method, we obtain accurate focal lengths, enabling us to leverage video for primary viewpoint texture information and mesh $\psi_{t}$ got before for geometric information from other viewpoints. Thus, we propose Gaussian-Mesh Contrastive Learning, as shown in Fig.~\ref{fig:3d2}. We randomly choose $N_{c'}$ camera poses $\left \{ c'_{i} \right \}_{N_{c'}}$, and each  corresponding focal is $f+\Delta  f$, $\Delta f$ is a slight and random perturbation.
Different from multiview DMs' productions, the rendered images of mesh $\psi_{t}$ are obtained from one entity, that naturally has multiview consistency. 
Besides, this method can provide references from any number of different viewpoints for training 4D Gaussian $\theta$, such density data can avoid artifacts in renderings.
The geometry alignment loss $\mathcal{L}_{\mathrm{GA}}$ can be summarized as:
\begin{align}  
    \mathcal{L}_{\mathrm{GA}} &= \sum_{i=1}^{N_{c'}} ( \mathcal{L}_{\mathrm{MSE}}(\mathrm{x}_{\theta_{t}}^{c'_{i}} , \mathrm{x}_{\psi_{t}}^{c'_{i}}) \notag \\ & +  \mathcal{L}_{\mathrm{Mask}}(\mathrm{x}_{\theta_{t}}^{c'_{i}} , \mathrm{x}_{\psi_{t}}^{c'_{i}})  + \lambda   \mathcal{L}_{\mathrm{LPIPS}}(\mathrm{x}_{\theta_{t}}^{c'_{i}} \mathrm{x}_{\psi_{t}}^{c'_{i}})).
\end{align}
Besides, we additionally introduce a T2I DM using the SDS method to enhance the control of the text prompt over the current object. Overall, the static alignment loss $\mathcal{L}_{\mathrm{static}}$ is denoted as:
\begin{equation}
    \mathcal{L}_{\mathrm{static}}=\mathcal{L}_{\mathrm{TA}}+\mathcal{L}_{\mathrm{GA}} + \mathcal{L}_{\mathrm{T2I}}.
    \label{eq:intra}
\end{equation}

\subsection{Dynamic Alignment Module} \label{sec:4D}
\textbf{Motion Alignment.}  With our focal alignment method, we can directly use the anchor video as the pixel-level alignment targets to provide motion guidance. Thus, we minimize the motion alignment loss $\mathcal{L}_{\mathrm{MA}}$ to inject dynamics:
\begin{equation}
\label{eq:mo}
\mathcal{L}_{\mathrm{MA}} = \frac{1}{T} \sum_{t=1}^{T}\sum_{H,W} || \mathrm{x}_{\theta_{t}}^{c} - I_\mathrm{vid}^{t} ||_{2}^{2},
\end{equation}
where $\mathrm{x}_{\theta_{t}}^{c}$ is the front-view renderings of 4D Gaussian at time $t$. $I_\mathrm{vid}^{t}$ is the corresponding frame of anchor video.

\begin{figure*}[ht]
    \centering
    \includegraphics[width=0.9\linewidth]{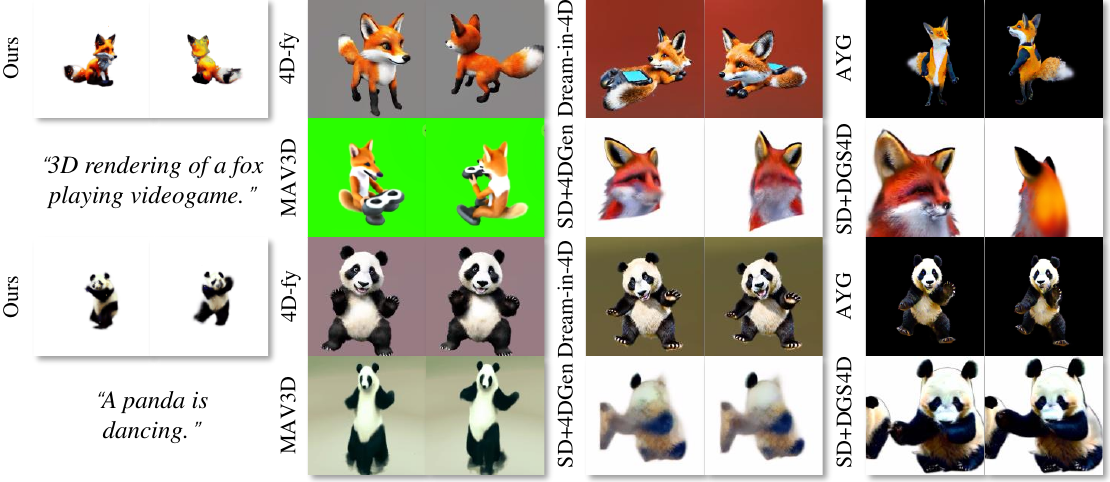}
    \caption{Comparison of PLA4D with text-to-4D and image-to-4D methods. Top: The pixel-level geometric priors provided by Gaussian-Mesh effectively help PLA4D avoid multi-face artifacts. The addition of the focal alignment module corrects the erroneous primary viewpoint projection relationships observed in image-to-4D methods. Bottom: With pixel-level alignment, PLA4D achieves the maximum motion range across 8-frame intervals, producing semantically coherent motion rather than pixel jittering. }
    \label{fig:compare}
\end{figure*}

\noindent \textbf{Time and Multiview Refinement.} Despite following the aforementioned technical steps to obtain a dynamic and geometrically reasonable 4D target, surface splitting may still occur. The Gaussian points with predicted locations are too far apart, and the scale cannot bridge the gap between these points. This indicates that some unfamiliar viewpoints still lack temporal continuity and geometric consistency. Thus, we propose the Time-Multiview (T-MV) Refinement, which uses the text prompt as a condition to optimize motion via video DM $\phi_{V}$, and the anchor video as a condition to optimize geometry via multiview DM $\phi_{MV}$, ensuring stable performance across multiple timestamps and random viewpoints. The $\mathcal{L}_{\mathrm{T\text{-}MV}}$ includes $\mathcal{L}_{\mathrm{Time}}$ and $\mathcal{L}_{\mathrm{MV}}$:
\begin{equation}
    \mathcal{L}_{\mathrm{Time}} = \frac{1}{\mathcal{T}}\sum_{t=1}^{\mathcal{T}}\sum_{H,W}w(\tau)||\epsilon_{\phi_{V}}(\alpha_{\tau}\mathrm{x}_{\theta_{t}}^{c'}+\sigma_{\tau} \epsilon ; \mathcal{C};\tau)-\epsilon||^{2}_{2},
\end{equation}
\begin{equation}
    \mathcal{L}_{\mathrm{MV}} = \frac{1}{N_{c'}}\sum_{i=1}^{N_{c'}}\sum_{H,W}w(\tau)||\epsilon_{\phi_{MV}}(\alpha_{\tau}\mathrm{x}_{\theta_{t}}^{c'_{i}}+\sigma_{\tau} \epsilon ; I_{\mathrm{vid}}^{t};\tau)-\epsilon||^{2}_{2},
\end{equation}
where $\tau$ is the timestep of DM, $w(\tau)$, $\alpha_{\tau}$ and $\sigma_{\tau}$ are parameters depends on the timestep $\tau$. Here, we can get $\mathcal{L}_{\mathrm{T\text{-}MV}} =  \mathcal{L}_{\mathrm{Time}} +  \mathcal{L}_{\mathrm{MV}}$.
In summary, we ultimately derive the dynamic alignment loss $\mathcal{L}_{dynamic}$:
\begin{equation}
    \mathcal{L}_{\mathrm{dynamic}} =\mathcal{L}_{\mathrm{MA}}  +\mathcal{L}_{\mathrm{T\text{-}MV}}.
    \label{eq:inter}
\end{equation}

\section{Experiments}

\begin{figure*}[t]
    \centering
    \includegraphics[width=\linewidth]{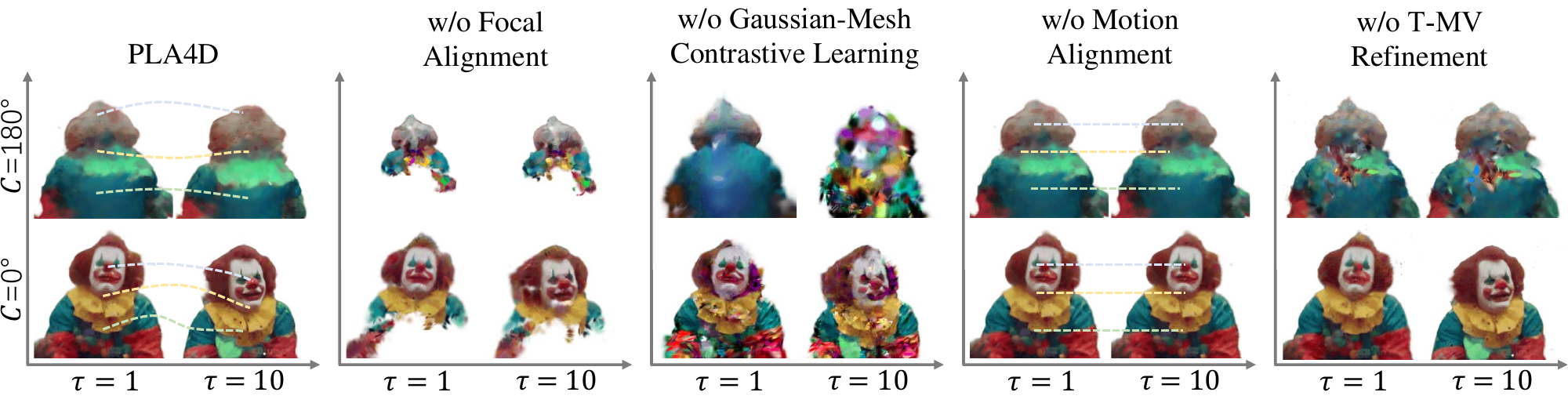}
    \caption{Ablation studies. If no focal alignment or Gaussian-Mesh contrastive learning, the 4D object loses its detailed texture and correct geometry. Without motion alignment, a 4D object degenerates into a static object. Absent T-MV refinement, the displacement of Gaussians causes surface tearing.}
    \label{fig:Ablation}
\end{figure*}

\textbf{Baselines.} 
For a comprehensive comparison, we evaluate our method alongside both text-to-4D methods~\cite{4d-fy,make-a-video-3d,AYG,4dgen,Dream-in-4D} and image-to-4D method~\cite{dreamgaussin4d}. For the image-to-4D methods, we use Stable Diffusion 2.1~\cite{sd2.1} with identical prompts to generate images, which are then used to generate 4D objects. Additionally, we compare methods based on both NeRF and Gaussian representations. For the closed-source methods MAV3D~\cite{make-a-video-3d} and AYG~\cite{AYG}, we perform comparisons using overlapping examples.

\noindent \textbf{Comparative Studies.} 
We present a large number of PLA4D-generated results in Fig.~\ref{fig:show}. Thanks to the pixel-level alignment methods, the 4D objects move beyond the rigid rendering style of previous 4D generation methods, exhibiting a stronger photorealistic style. Additionally, due to explicit motion guidance provided by the reference video, the target demonstrates detailed motion differences at each timestamp. Furthermore, with our proposed Gaussian-Mesh contrastive learning method, PLA4D's products also exhibit excellent geometric consistency. More importantly, each sample can be generated in just 15 minutes with 0.6K iterations.

Compared with other 4D generation methods, PLA4D demonstrates superior geometric structure, smooth motion, and semantic consistency, as shown in Fig.~\ref{fig:compare}. (1) Geometry: due to the implicit distillation of geometric priors in Dream-in-4D, the generated object suffers from the Janus-face problem. The same phenomenon can also be observed in the samples from MAV3D.
(2) Motion: due to the implicit distillation of motion priors, even when comparing the first and tenth frames, previous methods exhibit only small motion amplitudes, making it difficult to align with the motion described in the prompt.
(3) Semantic consistency: Although 4D-fy does not suffer from the Janus-face problem, its generated outputs exhibit semantic inconsistencies.
Due to the conflicts arising from the simultaneous optimization of multiple SDS objectives, where the optimization directions for geometry, motion, and semantics compete with each other, balancing these factors becomes challenging. PLA4D effectively alleviates this issue by employing pixel-level alignment.

The unified structure of NeRF with its MLP structure is not sensitive to each optimization step, allowing for better texture generation. In contrast, the Gaussian model optimizes each Gaussian point independently, making it more sensitive to each optimization step~\cite{GradeADreamer,t23dgs,dreamgaussin}. This structural difference introduces greater challenges in optimizing texture. However, PLA4D can still maintain high-quality texture by leveraging the T-MV refinement.

\begin{table}[!t]
    \centering
\resizebox{1\linewidth}{!}{
    \begin{tabular}{cccc}
    \toprule
         Methods & Representation & Generation Time & Iterations \\
    \midrule
         Animate124~\cite{Animate124}  & NeRF & - & 20K  \\
         4DGen~\cite{4dgen} & NeRF & 3.0 hr & 3K \\
         Consistent4D~\cite{Consistent4d}  & NeRF & 2.5 hr & 10K    \\
         DreamGaussian4D~\cite{dreamgaussin4d}  & Gaussians & 6.5 min & 0.7K  \\
    \midrule
         4D-fy~\cite{4d-fy}  & NeRF & 23 hr & 120K  \\
         Dream-in-4D~\cite{Dream-in-4D}   & NeRF & 10.5 hr & 20K   \\
         MAV3D~\cite{make-a-video-3d}  & NeRF & 6.5 hr & 12K   \\
         AYG~\cite{AYG}   & Gaussians  & \text{-} & 20K  \\
    \midrule
         PLA4D (ours)  & Gaussians  & 15 min & 0.6K  \\
    \bottomrule
    \end{tabular}}
    \caption{Speed comparison. The upper part presents image-to-4D methods, while the lower part collects text-to-4D methods..}
    \label{tab:time}
    \vspace{-10pt}
\end{table}

\begin{table}[!t]
    \centering
    \resizebox{1\linewidth}{!}{
    \begin{tabular}{c|c|c|c}
    \toprule
      Model   &  Motion   & Geometry   & Semantic consistency   \\
      \midrule
      4D-fy~\cite{4d-fy}   &  14.19 \%   & 21.10 \%  & 11.76 \% \\
      Dream-in-4D~\cite{Dream-in-4D} &  34.95 \%  & 27.68 \%  & 32.18 \%  \\
    \midrule
      PLA4D   &  50.86 \%  & 51.22 \%  & 56.06 \% \\
    \bottomrule
    \end{tabular}}
    \caption{User study. PLA4D receives the most praise from users for its consistency in motion, geometry, and semantics. }
    \label{tab:user}
    \vspace{-20pt}
\end{table}

\noindent \textbf{Ablation Study.} In Fig.~\ref{fig:Ablation}, we demonstrate the role of each module in PLA4D for text-to-4D generation.
Without the focal alignment method, using unmatched focal $f$, Gaussians can not learn the correct attributes of points to align to the generated frames. Both the geometry and motion of 4D objects are compromised.
Without Gaussian-Mesh contrastive learning, the geometry structure and texture in unknown views can not learned from multiview DM prior in such a short training time. 
Without motion alignment, the 4D object degrades into a static 3D object. Without T-MV refinement, dynamic multiview renderings of 4D objects result in surface cracks.

\noindent \textbf{Efficiency Study.} We compare the time overhead of multiple 4D generation methods proposed for image-to-4d and text-to-4d tasks, as shown in Tab.~\ref{tab:time}. It can be observed that previous 4D generation tasks overly rely on SDS, which requires extensive training (over 10K iterations) by implicitly aligning various diffusion models to generate 4D objects. In contrast, PLA4D uses explicit pixel-level alignment, resulting in better textures, geometry, and motion for 4D targets with significantly lower time overhead. 

\noindent \textbf{User Study.} To further evaluate the quality of our 4D generation objects, we conducted a user study on 30 participants. Specifically, we investigated users’ preference of 4D-fy~\cite{4d-fy}, Dream-in-4D~\cite{Dream-in-4D}, and our PLA4D in terms of motion, geometry, and semantic consistency. We didn't include MAV3D~\cite{make-a-video-3d} and AYG~\cite{AYG} because they are closed-source. As shown in Tab.~\ref{tab:user}, our PLA4D surpasses other comparison methods in all perspectives, indicating our superior performance on motion, geometry, and semantic consistency.

\noindent \textbf{Limitation.} PLA4D uses video as an anchor, relying on the performance of T2V DM. As the motion range of the text-driven generated video increases and the video duration extends, PLA4D will produce improved motion performance.

\section{Conclusion}
In this paper, we introduce PLA4D, a framework that leverages text-driven generated video as explicit pixel alignment targets for 4D generation, anchoring the rendering process conditioned by different DMs. We propose various modules to achieve such anchoring: we propose Gaussian-Mesh contrastive learning and focal alignment to ensure geometry consistency from the mesh and produce textures as detailed as those in the generated video frames. Additionally, we have developed a novel motion alignment method and T-MV refinement technology to optimize dynamic surfaces. Compared to existing methods, PLA4D effectively avoids Janus-face problem and generates 4D targets with accurate geometry and smooth motion in significantly less time. Furthermore, PLA4D is constructed entirely using existing open-source models, eliminating the need for pre-training any DMs. This flexible architecture allows the community to freely replace or upgrade components to achieve state-of-the-art performance. We aim for PLA4D to become an accessible, user-friendly, and promising tool for 4D digital content creation.

\clearpage
{
    \small
    \bibliographystyle{ieeenat_fullname}
    \bibliography{main}
}


\end{document}